%% file: main.tex
\documentclass[10pt,twocolumn,letterpaper]{article}

\usepackage{iccv}
\usepackage{times}
\usepackage{epsfig}
\usepackage{graphicx}
\usepackage{amsmath}
\usepackage{amssymb}
\usepackage{multirow}
\usepackage{caption}
\usepackage{graphicx, caption, subcaption}
\usepackage{cuted}
\usepackage{lscape}

\usepackage{fancyhdr}

\usepackage[pagebackref=true,breaklinks=true,letterpaper=true,colorlinks,bookmarks=false]{hyperref}

\iccvfinalcopy 

\def\iccvPaperID{4394} 

\begin{document}

\title{FairFace: Face Attribute Dataset for Balanced Race, Gender, and Age}

\author{Kimmo K\"arkk\"ainen\\
UCLA \\
{\tt\small kimmo@cs.ucla.edu }
\and
Jungseock Joo\\
UCLA \\ 
{\tt\small jjoo@comm.ucla.edu}
}

\maketitle


\begin{abstract}
Existing public face datasets are strongly biased toward Caucasian faces, and other races (e.g., Latino) are significantly underrepresented. This can lead to inconsistent model accuracy, limit the applicability of face analytic systems to non-White race groups, and adversely affect research findings based on such skewed data. To mitigate the race bias in these datasets, we construct a novel face image dataset, containing 108,501 images, with an emphasis of balanced race composition in the dataset. We define 7 race groups: White, Black, Indian, East Asian, Southeast Asian, Middle East, and Latino. Images were collected from the YFCC-100M Flickr dataset and labeled with race, gender, and age groups. Evaluations were performed on existing face attribute datasets as well as novel image datasets to measure generalization performance. We find that the model trained from our dataset is substantially more accurate on novel datasets and the accuracy is consistent between race and gender groups. 
The dataset will be released via \{\url{https://github.com/joojs/fairface}\}.
\end{abstract}

\input{000-intro.tex}

\input{001-related.tex}

\input{020-construct.tex}

\input{030-analyze.tex}

\input{060-bigtable.tex}

\input{050-conclusion.tex}

\section{Acknowledgement}
This work was supported by the National Science Foundation SMA-1831848, Hellman Fellowship, and UCLA Faculty Career Development Award. 

{\small
\bibliographystyle{ieee}
\bibliography{iccvbib}
}



\end{document}

%% file: 000-intro.tex

\section{Introduction}
To date, numerous large scale face image datasets~\cite{huang2008labeled,kumar2011describable,escalera2016chalearn,yi2014learning,liu2015faceattributes,joo2015automated,parkhi2015deep,yang2016wider,guo2016ms,kemelmacher2016megaface,Rothe-IJCV-2016,cao2018vggface2,merler2019diversity} have been proposed and fostered research and development for automated face detection~\cite{li2015convolutional,hu2017finding}, alignment~\cite{xiong2013supervised,ren2014face}, recognition~\cite{taigman2014deepface,schroff2015facenet}, generation~\cite{yan2016attribute2image,bao2017cvae,karras2018style,thomas2018persuasive}, modification~\cite{antipov2017face,lample2017fader,he2017arbitrary}, and attribute classification~\cite{kumar2011describable,liu2015faceattributes}. These systems have been successfully translated into many areas including security, medicine, education, and social sciences. 

\begin{table*}[th!]
\small
\caption{Statistics of Face Attribute Datasets}
\begin{tabular}{|c|c|c|c|c|c|c|c|c|c|c|c|c|c|c|}
\cline{7-14}
\multicolumn{6}{c}{}      & \multicolumn{8}{|c|}{Race Annotation} \\ \cline{1-14}
\multirow{2}*{Name} & 
\multirow{2}*{Source} &
\multirow{2}*{\shortstack{\# of \\ faces}} &
\multirow{2}*{\shortstack{In-the-\\wild?}} &
\multirow{2}*{Age} &
\multirow{2}*{Gender}  & \multicolumn{2}{c|}{White*} & \multicolumn{2}{c|}{Asian*} & \multirow{2}*{\shortstack{Bla-\\ck}} & \multirow{2}*{\shortstack{Ind-\\ian}} & \multirow{2}*{\shortstack{Lat-\\ino}} & \multirow{2}*{\shortstack{Balan-\\ced?}}  \\ \cline{7-10}
& &&&& & W & ME & E & SE &  &  &  &  \\ \hline
\hline


\multicolumn{1}{|c|}{PPB \cite{buolamwini2018gender}} & \begin{tabular}{@{}c@{}}Gov. Official \\ Profiles\end{tabular}   &  1K   &       &  \checkmark & \checkmark & \multicolumn{8}{c|}{**Skin color prediction}  \\ \hline

\multicolumn{1}{|c|}{MORPH \cite{ricanek2006morph}} & Public Data &    55K   &       &  \checkmark & \checkmark & \multicolumn{2}{c|}{merged} &  & & \checkmark &   & \checkmark & no \\ \hline

\multicolumn{1}{|c|}{PubFig \cite{kumar2011describable}} & Celebrity &    13K   & \checkmark      &  \multicolumn{9}{c|}{Model generated predictions}   & no \\ \hline

\multicolumn{1}{|c|}{IMDB-WIKI \cite{Rothe-IJCV-2016}} & IMDB, WIKI &   500K    &   \checkmark    &  \checkmark &  \checkmark &  &  &  &  & & & & no \\ \hline

\multicolumn{1}{|c|}{FotW \cite{escalera2016chalearn}} & Flickr  &  25K   &   \checkmark    &  \checkmark & \checkmark &  &  &  & & &  &  & yes \\ \hline

\multicolumn{1}{|c|}{CACD \cite{chen2015face}} & celebrity &    160K   &   \checkmark    &  \checkmark &  &  &  &  &  &  & & & no \\ \hline

\multicolumn{1}{|c|}{DiF \cite{merler2019diversity}} & Flickr  &  1M   &   \checkmark    &  \checkmark & \checkmark & \multicolumn{8}{c|}{**Skin color prediction}  \\ \hline

\multicolumn{1}{|c|}{\dag CelebA \cite{liu2015faceattributes}} &  \begin{tabular}{@{}c@{}}CelebFace~\cite{sun2013hybrid,sun2014deep}\\ LFW \cite{LFWTech}\end{tabular}  &  200K   &   \checkmark    &  \checkmark  & \checkmark &  &  &  &  & & & & no \\ \hline

\multicolumn{1}{|c|}{LFW+ \cite{han2018heterogeneous}} & \begin{tabular}{@{}c@{}}LFW \cite{LFWTech}\\ (Newspapers)\end{tabular} &   15K    &   \checkmark    &  \checkmark & \checkmark & \multicolumn{2}{c|}{merged} & \multicolumn{3}{c|}{merged} &  &  & no \\ \hline

\multicolumn{1}{|c|}{\dag LFWA+ \cite{liu2015faceattributes}} & \begin{tabular}{@{}c@{}}LFW \cite{LFWTech}\\ (Newspapers)\end{tabular} &  13K     &   \checkmark    &  & \checkmark & \multicolumn{2}{c|}{merged} & \multicolumn{2}{c|}{merged} & \checkmark & \checkmark  &  & no \\ \hline

\multicolumn{1}{|c|}{\dag UTKFace \cite{zhang2017age}} & \begin{tabular}{@{}c@{}}MORPH, CACD \\ Web\end{tabular}  &  20K   &   \checkmark    &  \checkmark & \checkmark & \multicolumn{2}{c|}{merged}  & \multicolumn{2}{c|}{merged} & \checkmark & \checkmark &  & yes \\ \hline
\hline

\multicolumn{1}{|c|}{\textbf{FairFace (Ours)}} & \begin{tabular}{@{}c@{}}Flickr, Twitter \\ Newspapers, Web\end{tabular}  &  108K   &   \checkmark    &  \checkmark & \checkmark & \checkmark &   \checkmark    &  \checkmark & \checkmark & \checkmark & \checkmark & \checkmark & yes  \\ \hline

\multicolumn{14}{r}{*FairFace (Ours) also defines East (E) Asian, Southeast (SE) Asian, Middle Eastern (ME), and Western (W) White. } \\ 
\multicolumn{14}{r}{**PPB and DiF do not provide race annotations but skin color annotated or automatically computed as a proxy to race. } \\
\multicolumn{14}{r}{\dag denotes datasets used in our experiments. }
\end{tabular}
\label{table:stat}
\end{table*}

Despite the sheer amount of available data, existing public face datasets are strongly biased toward Caucasian faces, and other races (e.g., Latino) are significantly underrepresented. A recent study shows that most existing large scale face databases are biased towards ``lighter skin'' faces (around 80\%), \eg, White, compared to ``darker'' faces, \eg, Black~\cite{merler2019diversity}. This means the model may not apply to some subpopulations and its results may not be compared across different groups without calibration. Biased data will produce biased models trained from it. This will raise ethical concerns about fairness of automated systems, which has emerged as a critical topic of study in the recent machine learning and AI literature~\cite{hardt2016equality,corbett2017algorithmic}. 

For example, several commercial computer vision systems (Microsoft, IBM, Face++) have been criticized due to their asymmetric accuracy across sub-demographics in recent  studies~\cite{buolamwini2018gender,raji2019actionable}. These studies found that the commercial face  gender classification systems all perform better on male and on light faces. This can be caused by the biases in their training data. Various unwanted biases in image datasets can easily occur due to biased selection, capture, and negative sets~\cite{torralba2011unbiased}. Most public large scale face datasets have been collected from popular online media -- newspapers, Wikipedia, or webs search-- and these platforms are more frequently used by or showing White people. 

To mitigate the race bias in the existing face datasets, we propose a novel face dataset with an emphasis of balanced race composition. Our dataset contains 108,501 facial images collected primarily from the YFCC-100M Flickr dataset~\cite{thomee59yfcc100m}, which can be freely shared for a research purpose, and also includes examples from other sources such as Twitter and online newspaper outlets. We define 7 race groups: White, Black, Indian, East Asian, Southeast Asian, Middle East, and Latino. Our dataset is well-balanced on these 7 groups (See Figure~\ref{fig:grid} and \ref{fig:race-comp})

Our paper make three main contributions. 
First, we emprically show that existing face attribute datasets and models learned from them do not generalize well to unseen data in which more non-White faces are present. Second, we show that our new dataset perform better on the novel data, not only on average, but also across racial groups, \ie more consistent. Third, to the best of our knowledge, our dataset is the first large scale face attribute dataset in the wild which includes Latino and Middle Eastern and differentiates East Asian and South East Asian. Computer vision has been rapidly transferred into other fields such as economics or social sciences, where researchers want to analyze different demographics using image data. The inclusion of major racial groups, which have been missing in existing datasets, therefore significantly enlarges the applicability of computer vision methods to these fields.   


%% file: 001-related.tex

\section{Related Work}
\subsection{Face Attribute Recognition}
Face attribute recognition is a task to classify various human attributes such as gender, race, age, emotions, expressions or other facial traits from facial appearance~\cite{kumar2011describable,joo2013human,zhang2015learning,liu2015faceattributes}.  While there have been many techniques developed for the task, we mainly review  datasets which are the main concern of this paper. 


Table~\ref{table:stat} summarizes the statistics of existing large scale face attribute datasets and our new dataset. The is not an exhaustive list but we focus on \textbf{public} and \textbf{in-the-wild} datasets on gender, race, and age. As stated earlier, most of these datasets were constructed from online sources, which are typically dominated by the White race. 

Face attribute recognition has been applied as a sub-component to other computer vision systems. For example, Kumar et al.~\cite{kumar2011describable} used facial attributes such as gender, race, hair style, expressions, and accessories as features for face verification, as the attributes characterize individual traits. Attributes are also widely used for person re-identification in images or videos, combining features from human face and body appearance~\cite{layne2012person,li2015clothing,su2018multi}, especially effective when faces are not fully visible or too small. These systems have applications in security such as authentication for electronic devices (\eg, unlocking smartphones) or monitoring surveillance CCTVs~\cite{grgic2011scface}. 

It is imperative to ensure that these systems perform evenly well on difference gender and race groups. Failing to do so can be detrimental to the reputations of individual service providers and the public trust about the machine learning and computer vision research community. Most notable incidents regarding the racial bias include Google Photos recognizing African American faces as Gorilla and Nikon's digital cameras prompting a message asking ``did someone blink?'' to Asian users~\cite{zhang_2015}. These incidents, regardless of whether the models were trained improperly or how much they actually affected the users, often result in the termination of the service or features (\eg, dropping sensitive output categories). For the reason, most commercial service providers have stopped providing a race classifier. 

Face attribute recognition is also widely used for demographic survey performed in marketing or social science research, aimed at understanding human social behaviors and their relations to demographic backgrounds of individuals. Using off-the-shelf tools~\cite{amos2016openface,baltrusaitis2018openface} and commercial services, social scientists, who traditionally didn't use images, begun to use images of people to infer their demographic attributes and analyze their behaviors in many studies. Notable examples are demographic analyses of social media users using their  photographs~\cite{chakraborty2017makes,reis2017demographics,won2017protest,xi2019understanding,wang2017polarized}. 
The cost of unfair classification is huge as it can over- or under-estimate specific sub-populations in their analysis, which may have policy implications. 


\subsection{Fair Classification and Dataset Bias}
Researchers in AI and machine learning have increasingly paid attention to algorithmic fairness and dataset and model biases~\cite{zemel2013learning,corbett2017algorithmic,zou2018ai,zhang2018mitigating}. There exist many different definitions of fairness used in the literature~\cite{verma2018fairness}. In this paper, we focus on balanced accuracy--whether the attribute classification accuracy is independent of race and gender. More generally, research in fairness is concerned with a model's ability to produce fair outcomes (\eg, loan approval) independent of protected or sensitive attributes such as race or gender. 


Studies in algorithmic fairness have focused on either 1) discovering (auditing) existing bias in datasets or systems~\cite{shankar2017no,buolamwini2018gender,kiritchenko2018examining}, 2) making a better dataset~\cite{merler2019diversity,Alvi_2018_ECCV_Workshops}, or 3) designing a better algorithm or model~\cite{Das_2018_ECCV_Workshops,Alvi_2018_ECCV_Workshops,ryu2017inclusivefacenet,zemel2013learning,zafar2017fairness}. Our paper falls into the first two categories. 

In computer vision, it has been shown that popular large scale image datasets such as Imagenet are biased in terms of the origin of images (45\% were from the U.S.) \cite{suresh2018learning} or the underlying association between scene and race~\cite{stock2018convnets}.  Can we make a perfectly balanced dataset? It is ``infeasible to balance across all possible co-occurrences'' of attributes~\cite{hendricks2018women}. This is possible in a lab-controlled setting, but not in a dataset ``in-the-wild''. 

Therefore, the contribution of our paper is to mitigate, not entirely solve, the current limitation and biases of existing databases by collecting more diverse face images from non-White race groups. We empirically show this significantly improves the generalization performance to novel image datasets whose racial compositions are not dominated by the White race. Furthermore, as shown in Table~\ref{table:stat}, our dataset is the first large scale in-the-wild face image dataset which includes Southeast Asian and Middle Eastern races. While their faces share similarity with East Asian and White groups, we argue that not having these major race groups in datasets is a strong form of discrimination. 


%% file: 020-construct.tex
\begin{figure}
  \centering
      \includegraphics[width=0.5\textwidth]{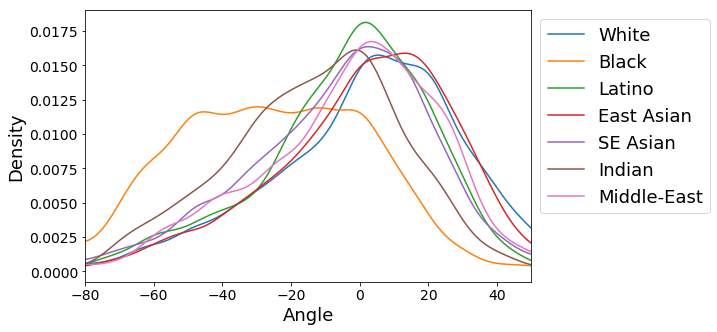}
  \caption{Individual Typology Angle (ITA), \ie, skin color, distribution of different races measured in our dataset.}
\label{fig:skincolor}
\end{figure}

\section{Dataset Construction}
\subsection{Race Taxonomy}

In this study, we define 7 race groups: White, Black, Indian, East Asian, Southeast Asian, Middle East, and Latino.
Race and ethnicity are different categorizations of humans. Race is defined based on physical trait and ethnicity is based on cultural similarities~\cite{schaefer2008encyclopedia}. For example, Asian immigrants in Latin America can be of Latino ethnicity. In practice, these two terms are often used interchangeably. Race is not a discrete concept and needs to be clearly defined before data collection. 

We first adopted a commonly accepted race classification from the U.S. Census Bureau (White, Black, Asian, Hawaiian and Pacific Islanders, Native Americans, and Latino). Latino is often treated as an ethnicity, but we consider Latino a race, which can be judged from the facial appearance. We then further divided subgroups such as Middle Eastern, East Asian, Southeast Asian, and Indian, as they look clearly distinct. During the data collection, we found very few examples for Hawaiian and Pacific Islanders and Native Americans and discarded these categories. All the experiments conducted in this paper were therefore based on 7 race classification.

\begin{figure}
    \centering
      \includegraphics[width=0.5\textwidth]{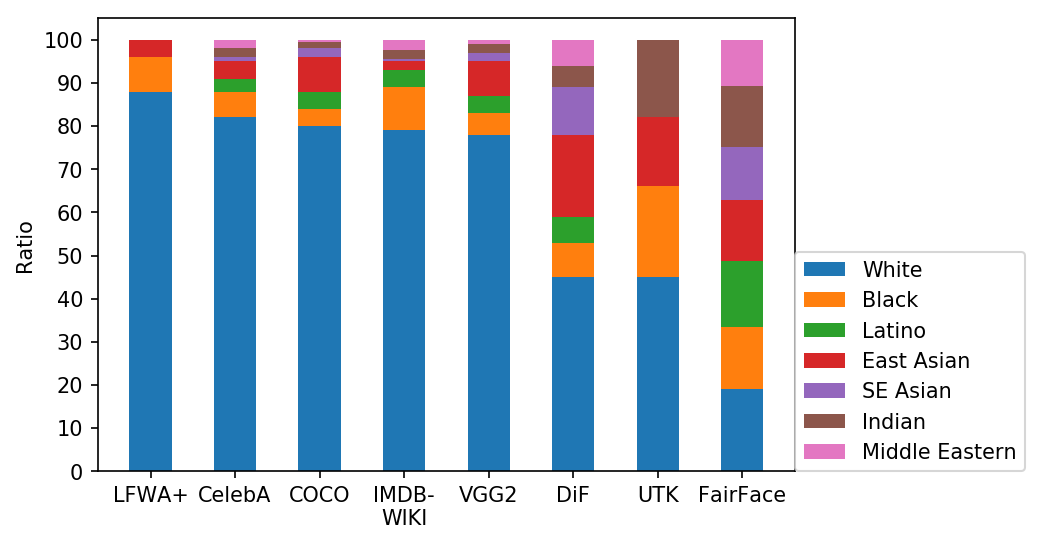}
  \caption{Racial compositions in face datasets.}
\label{fig:race-comp}
\end{figure}

\begin{figure*}
    \centering
    \begin{subfigure}[t]{0.3\textwidth}
    \centering
      \includegraphics[width=1\textwidth]{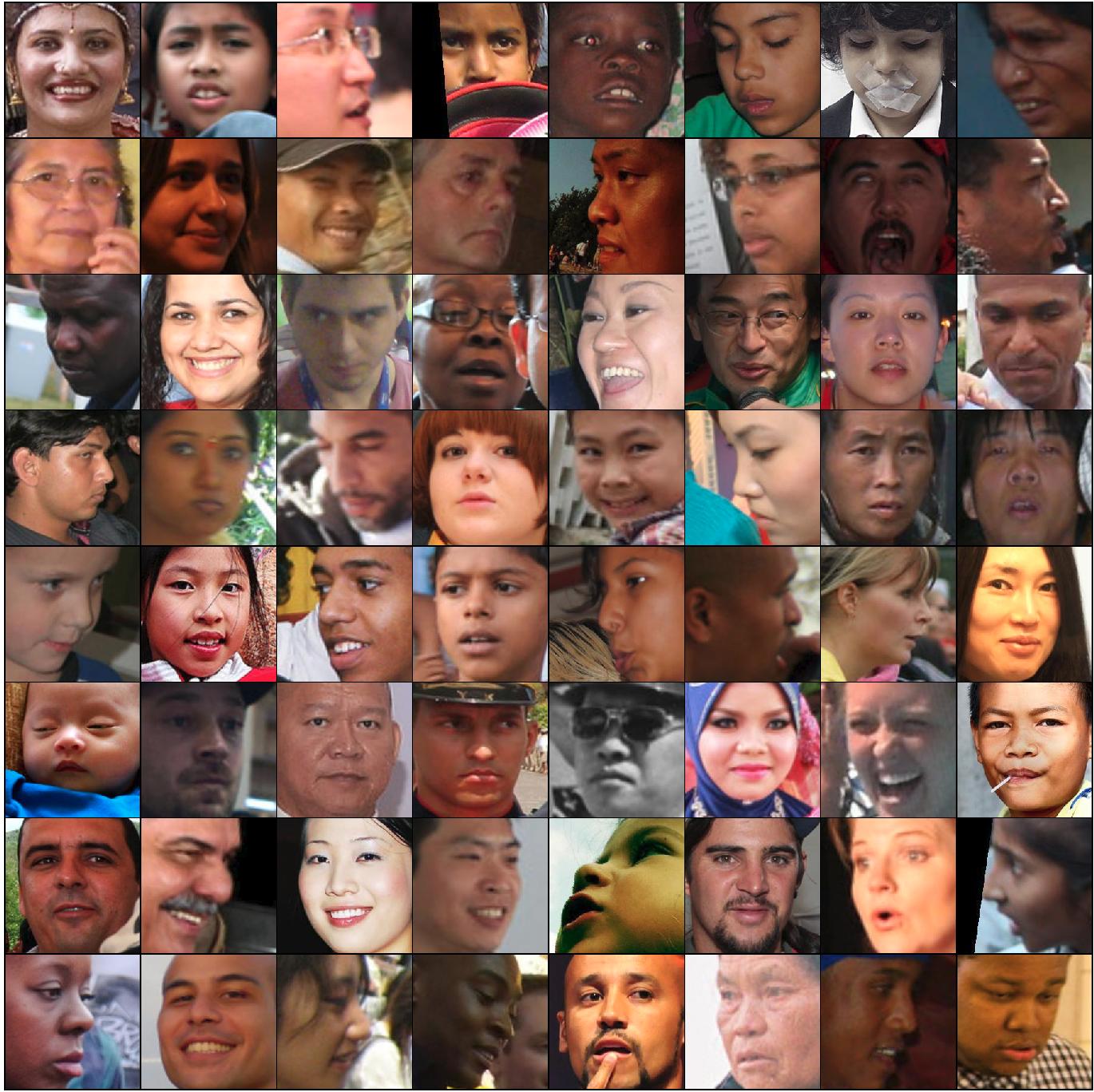}
      \caption{FairFace}
     \end{subfigure}    ~~~
    \begin{subfigure}[t]{0.3\textwidth}
    \centering
      \includegraphics[width=1\textwidth]{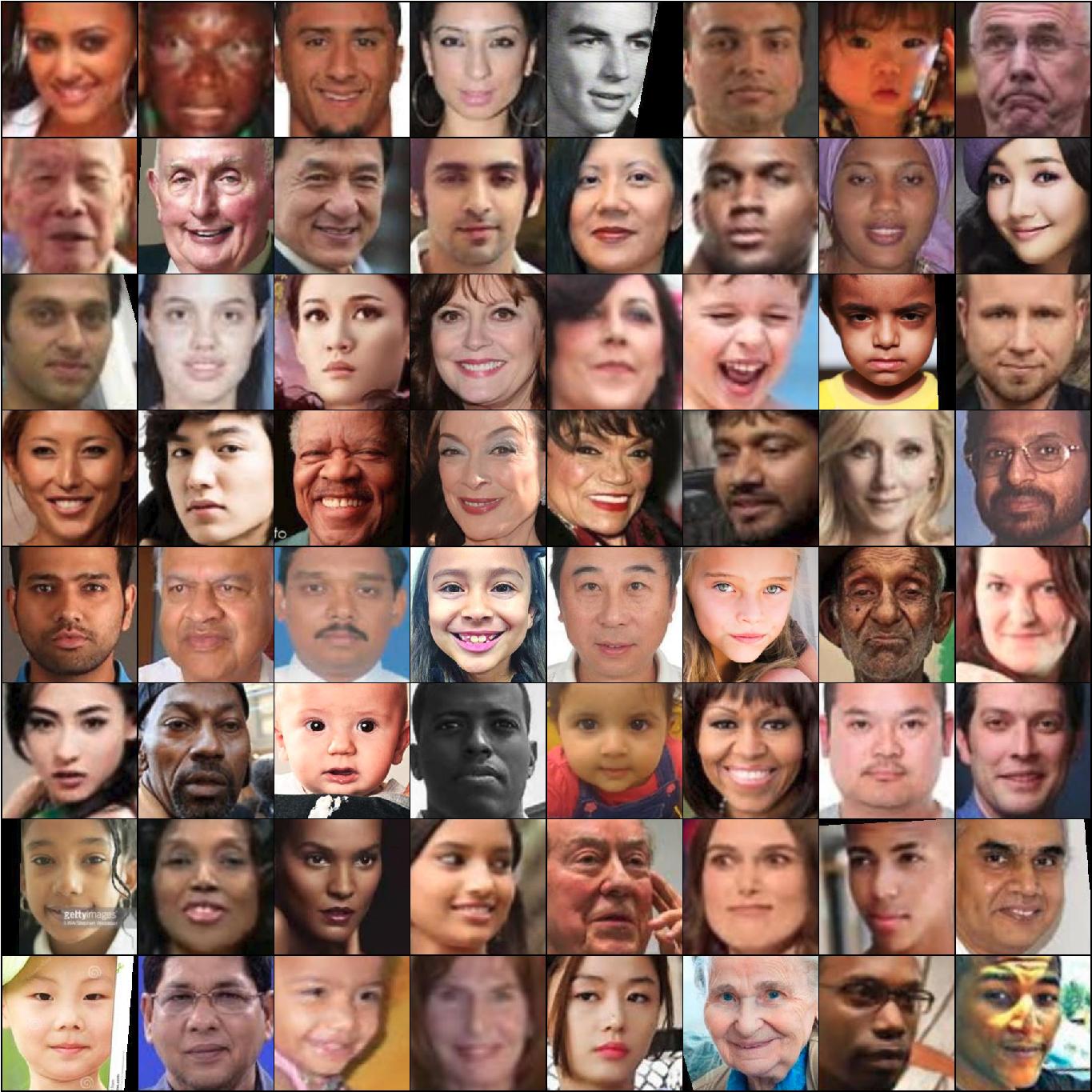}
      \caption{UTKFace}
     \end{subfigure}    \\
    \begin{subfigure}[t]{0.3\textwidth}
    \centering
      \includegraphics[width=1\textwidth]{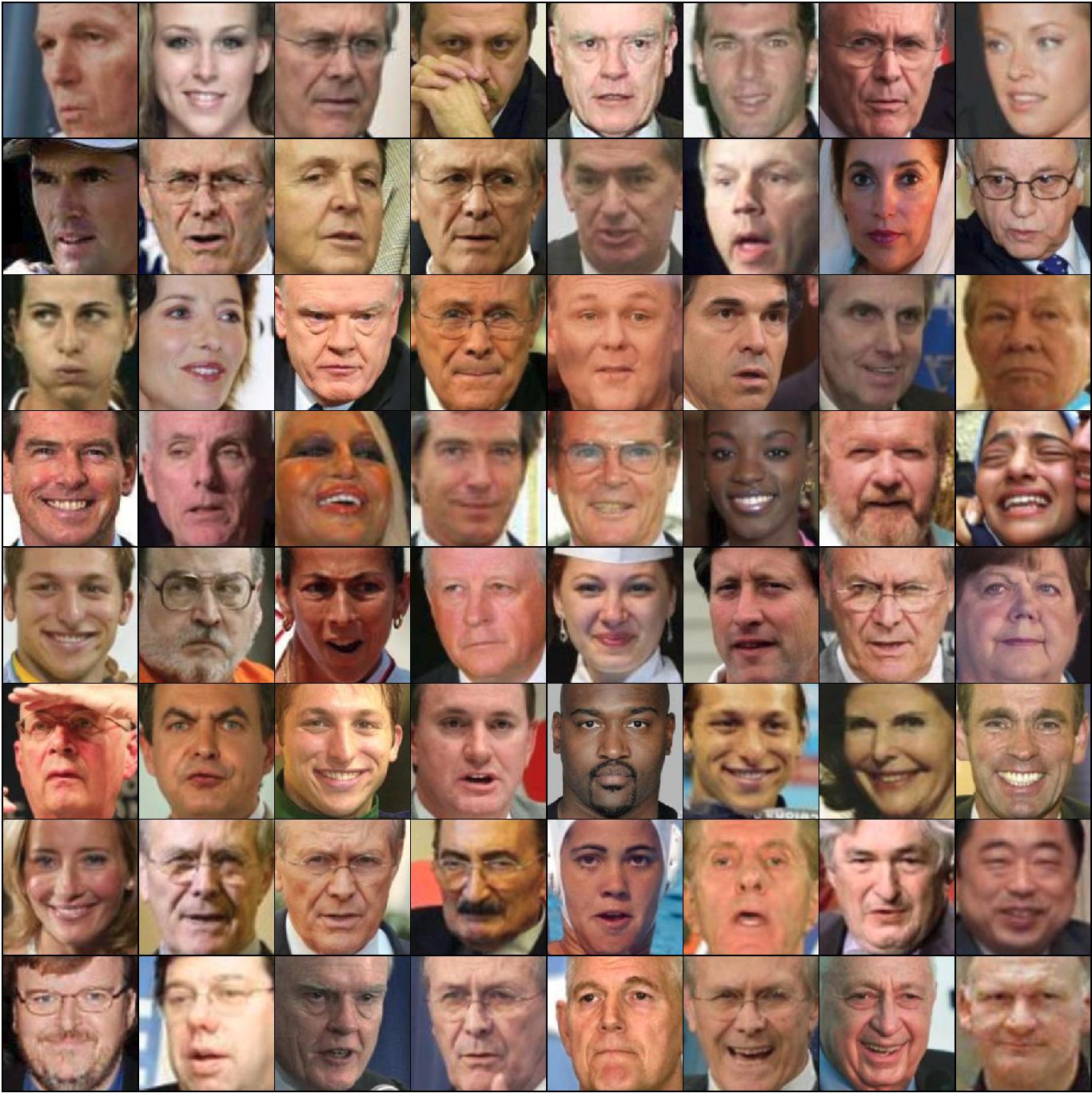}
      \caption{LFWA+}
     \end{subfigure}    ~~~
    \begin{subfigure}[t]{0.3\textwidth}
    \centering
      \includegraphics[width=1\textwidth]{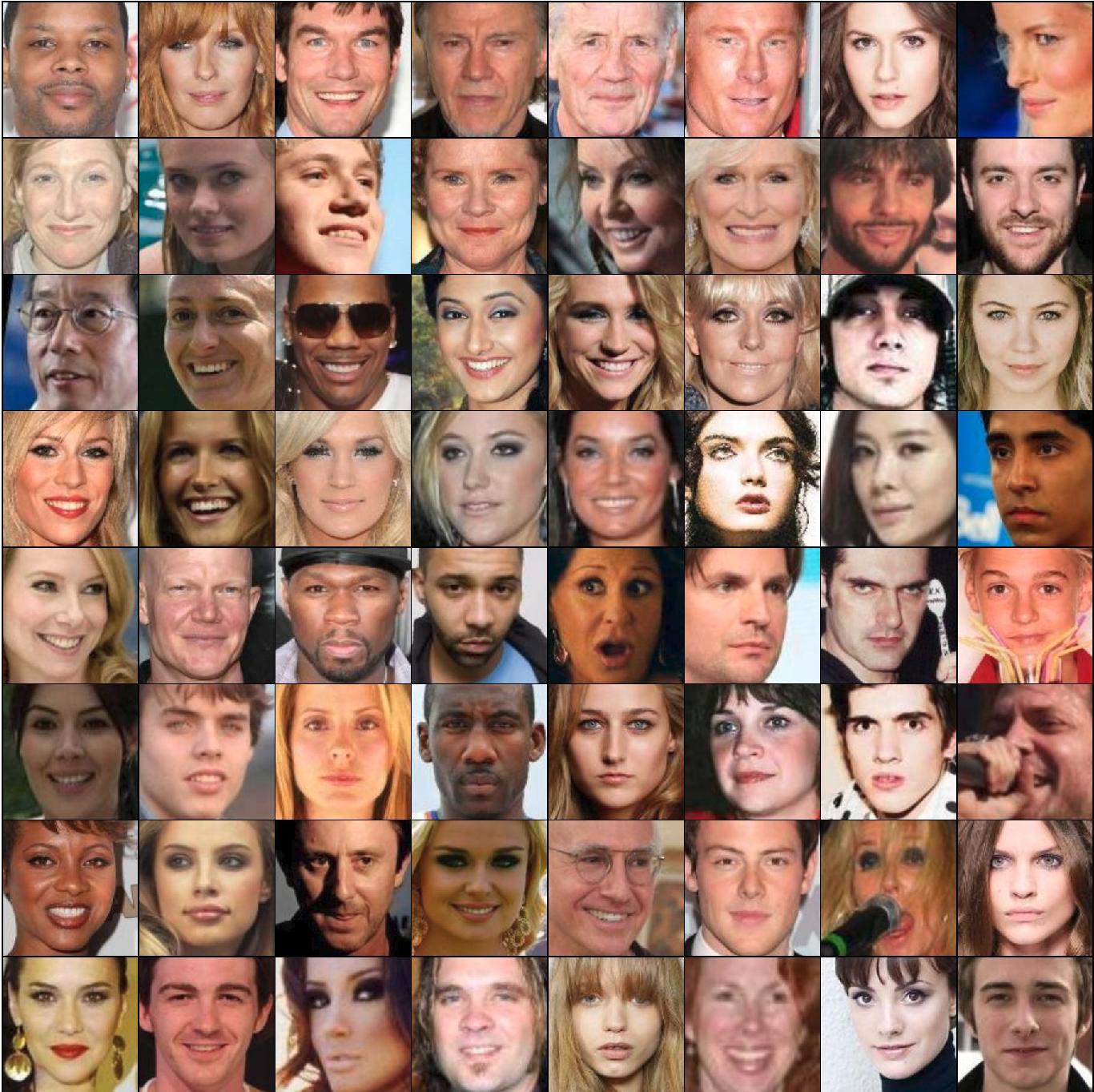}
      \caption{CelebA}
     \end{subfigure}    
  \caption{Random samples from face attribute datasets.}
\label{fig:grid}
\end{figure*}




A few recent studies \cite{buolamwini2018gender,merler2019diversity} use skin color as a proxy to racial or ethnicity grouping. While skin color can be easily computed without subjective annotations, it has  limitations. First, skin color is heavily affected by illumination and light conditions. The Pilot Parliaments Benchmark (PPB) dataset \cite{buolamwini2018gender} only used profile photographs of government officials taken in well controlled lighting, which makes it non-in-the-wild. Second, within-group variations of skin color are huge. Even same individuals can show different skin colors over time. Third, most importantly, race is a multidimensional concept whereas skin color (\ie, brightness) is one dimensional. Figure~\ref{fig:skincolor} shows the distributions of the skin color of multiple race groups, measured by Individual Typology Angle (ITA)~\cite{wilkes2015fitzpatrick}. As clearly shown here, the skin color provides no information to  differentiate many groups such as East Asian and White. 
Therefore, we explicitly use race and annotate the physical race by human annotators' judgments. 


\subsection{Image Collection and Annotation}
Many existing face datasets have relied on photographs of public figures such as politicians or celebrities~\cite{kumar2011describable,LFWTech,joo2015automated,Rothe-IJCV-2016,liu2015faceattributes}. Despite the easiness of collecting images and ground truth attributes, the selection of these populations may be biased. For examples, politicians may be older and actors may be more attractive than typical faces. 
Their images are usually taken by professional photographers in limited situations, leading to the quality bias. Some datasets were collected via web search using keywords such as ``Asian boy'' \cite{zhang2017age}. These queries may return only stereotypical faces or prioritize celebrities in those categories rather than diverse individuals among general public. 



Our goal is to minimize the selection bias introduced by such filtering and maximize the diversity and coverage of the dataaset. We started from a huge public image dataset, Yahoo YFCC100M dataset~\cite{thomee59yfcc100m}, and detected faces from the images without any preselection. A recent work also used the same dataset to construct a huge unfiltered face dataset (Diversity in Face, DiF) \cite{merler2019diversity}. Our dataset is smaller but more balanced on race (See Figure~\ref{fig:race-comp}). 

Specifically, we incrementally increased the dataset size. We first detected and annotated 7,125 faces randomly sampled from the entire YFCC100M dataset ignoring the locations of images. After obtaining annotations on this initial set, we estimated demographic compositions of each country. Based on this statistic, we adaptively adjusted the number of images for each country sampled from the dataset such that the dataset is not dominated by the White race. Consequently, we excluded the U.S. and European countries in the later stage of data collection after we sampled enough White faces from those countries. 
The minimum size of a detected face was set to 50 by 50 pixels. This is a relatively smaller size compared to other datasets, but we find the attributes are still recognizable and these examples can actually make the classifiers more robust against noisy data. We only used images with ``Attribution'' and ``Share Alike'' Creative Commons licenses, which allow derivative work and commercial usages.

We used Amazon Mechanical Turk to verify the race, gender and age group for each face. We assigned three workers for each image. If two or three workers agreed on their judgements, we took the values as ground-truth. If all three workers produced different responses, we republished the image to another 3 workers and subsequently discarded the image if the new annotators did not agree. 
These annotations at this stage were still noisy. We further refined the annotations by training a model from the initial ground truth annotations and applying back to the dataset. We then manually re-verified the annotations for images whose annotations differ from model predictions. 


%% file: 030-analyze.tex
\section{Experiments}
\subsection{Measuring Bias in Datasets}
We first measure how skewed each dataset is in terms of its race composition. 
For the datasets with race annotations, we use the reported statistics. For the other datasets, we annotated the race labels for 3,000 random samples drawn from each dataset. See Figure~\ref{fig:race-comp} for the result. As expected, most existing face attribute datasets, especially the ones focusing on celebrities or politicians, are biased toward the White race. Unlike race, we find that most datasets are relatively more balanced on gender ranging from 40\%-60\% male ratio. 

\subsection{Model and Cross-Dataset Performance}
To compare model performance of different datasets, we used an identical model architecture, ResNet-34~\cite{he2016deep}, to be trained from each dataset. We used ADAM optimization \cite{kingma2014adam} with a learning rate of 0.0001. Given an image, we detected faces using the dlib\footnote{dlib.net}'s CNN-based face detector~\cite{king2015max} and ran the attribute classifier on each face. The experiment was done in pyTorch. 

Throughout the evaluations, we compare our dataset with three other datasets: UTKFace~\cite{zhang2017age}, LFWA+, and CelebA~\cite{liu2015faceattributes}. Both UTKFace and LFWA+ have race annotations, and thus, are suitable for comparison with our dataset. CelebA does not have race annotations, so we only use it for gender classification. See Table~\ref{table:stat} for more detailed dataset characteristics.

\begin{figure*}
    \centering
    \begin{subfigure}[t]{0.33\textwidth}
    \centering
      \includegraphics[width=1\textwidth]{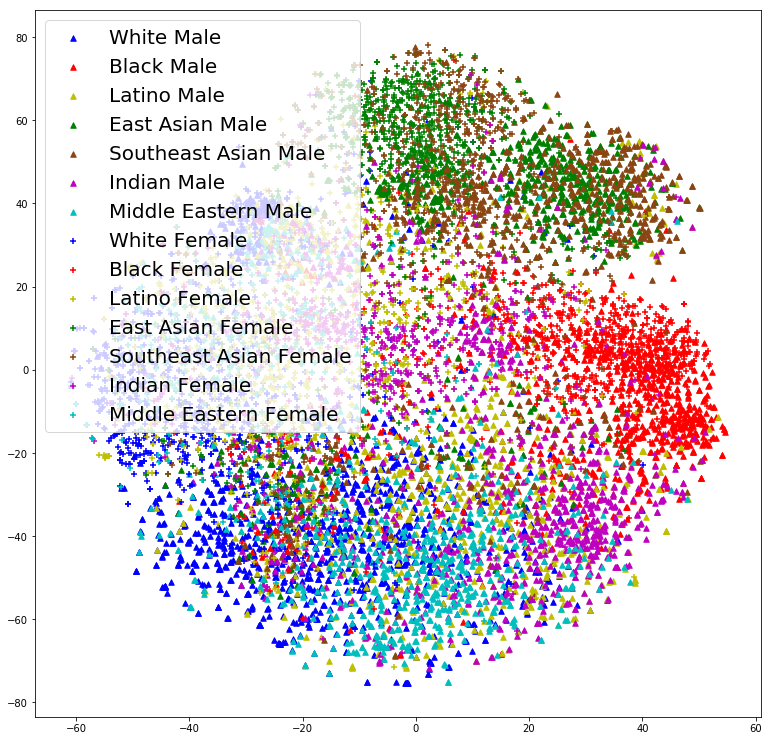}
      \caption{FairFace}
     \end{subfigure}
    \begin{subfigure}[t]{0.33\textwidth}  \centering
      \includegraphics[width=1\textwidth]{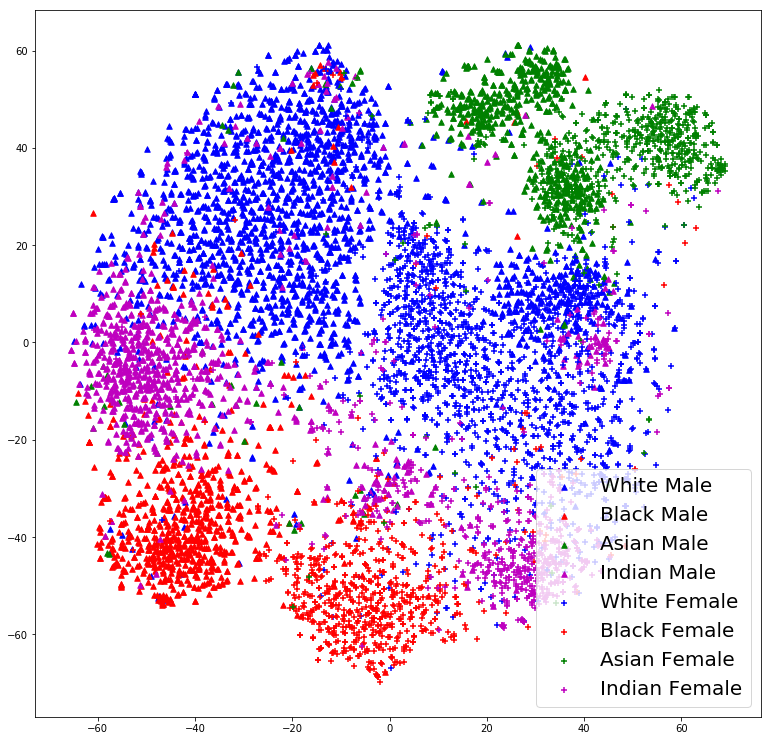}
      \caption{UTKFace}
     \end{subfigure}
    \begin{subfigure}[t]{0.33\textwidth}  \centering
      \includegraphics[width=1\textwidth]{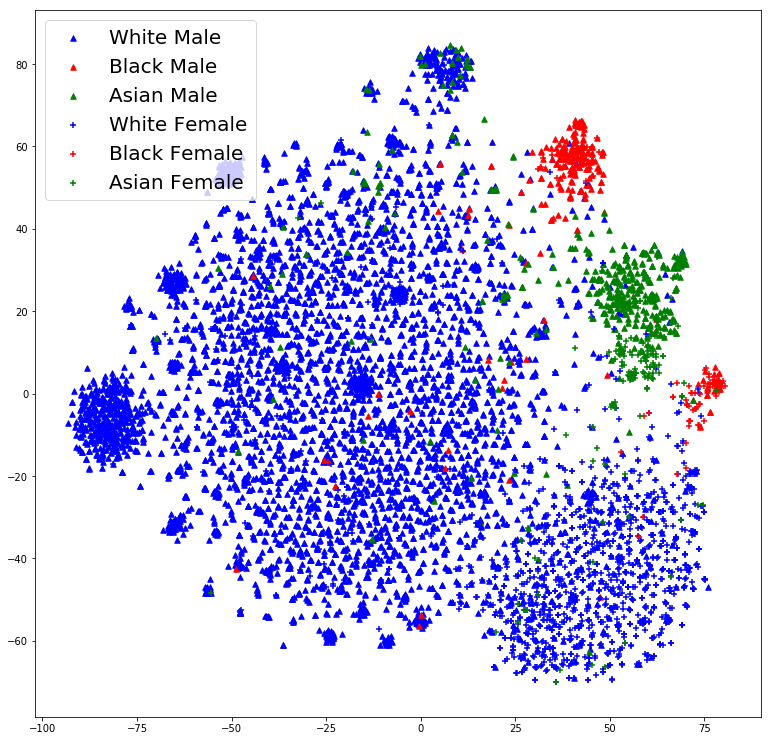}
      \caption{LFWA+}
     \end{subfigure}
  \caption{t-SNE visualizations \cite{maaten2008visualizing} of faces in datasets.} 
\label{fig:tsne}
\end{figure*}

Using models trained from these datasets, we first performed cross-dataset classifications, by alternating training sets and test sets. Note that FairFace is the only dataset with 6 races. To make it compatible with other datasets, we merged our fine racial groups when tested on other datasets. CelebA does not have race annotations but was included for gender classification. 

Tables~\ref{table:cross_white} and \ref{table:cross_non_white} show the classification results for race, gender, and age on the datasets across subpopulations. As expected, each model tends to perform better on the same dataset on which it was trained. However, the accuracy of our model was highest on some variables on the LFWA+ dataset and also very close to the leader in other cases. This is partly because LFWA+ is the most biased dataset and ours is the most diverse, and thus more generalizable dataset.

\subsection{Generalization Performance}
\subsubsection{Datasets}
To test the generalization performance of the models, we consider three novel datasets. Note that these datasets were collected from completely different sources than our data from Flickr and not used in training. Since we want to measure the effectiveness of the model on diverse races, we chose the test datasets that contain people in different locations as follows.

\textbf{Geo-tagged Tweets.} 
First we consider images uploaded by Twitter users whose locations are identified by geo-tags (longitude and latitude), provided by \cite{steinert2018twitter}. From this set, we chose four countries (France, Iraq, Philippines, and Venezuela) and randomly sampled 5,000 faces. 

\textbf{Media Photographs.} 
Next, we also use photographs posted by 500 online professional media outlets. Specifically, we use a public dataset of tweet IDs \cite{DVN/2FIFLH_2017} posted by 4,000 known media accounts, \eg, $@$nytimes. Note that although we use Twitter to access the photographs, these tweets are simply external links to pages in the main newspaper sites. Therefore this data is considered as media photographs and different from general tweet images mostly uploaded by ordinary users. We randomly sampled 8,000 faces from the set.

\textbf{Protest Dataset.} 
Lastly, we also use a public image dataset collected for a recent protest activity study \cite{won2017protest}. The authors collected the majority of data from Google Image search by using keywords such as ``Venezuela protest'' or ``football game'' (for hard negatives). The dataset exhibits a wide range of diverse race and gender groups engaging in different activities in various countries. We randomly sampled 8,000 faces from the set. 

These faces were annotated for gender, race, and age by Amazon Mechanical Turk workers. 

\subsubsection{Result}
Table~\ref{table:validation} shows the classification accuracy of different models. Because our dataset is larger than LFWA+ and UTKFace, we report the three variants of the FairFace model by limiting the size of a training set (9k, 18k, and Full) for fair comparisons. 


\textbf{Improved Accuracy. }
As clearly shown in the result, the model trained by FairFace outperforms all the other models for race, gender, and age, on the novel datasets, which have never been used in training and also come from different data sources. The models trained with fewer training images (9k and 18k) still outperform other datasets including CelebA which is larger than FairFace. This suggests that the dataset size is not the only reason for the performance improvement. 

\textbf{Balanced Accuracy. }
Our model also produces more consistent results -- for race, gender, age classification -- across different race groups compared to other datasets. We measure the model consistency by standard deviations of classification accuracy measured on different sub-populations, as shown in Table~\ref{table:mean_var}. More formally, one can consider conditional use accuracy equality~\cite{berk2018fairness} or equalized odds~\cite{hardt2016equality} as the measure of fair classification. For gender classification:

\begin{align}
 P(\widehat{Y} = i | Y = i, A = j) = P(\widehat{Y} = i | Y = i, A = k), \nonumber  \\
 i \in \{\textrm{male, female}\},  \forall j, k \in \mathrm{D}, 
\end{align}
where $\widehat{Y}$ is the predicted gender, $Y$ is the true gender, A refers to the demographic group, and $\mathrm{D}$ is the set of different demographic groups being considered (\ie race). When we consider different gender groups for $A$, this needs to be modified to measure accuracy equality~\cite{berk2018fairness}: 
\begin{align}
 P(\widehat{Y} = Y | A = j) = P(\widehat{Y} = Y | A = k), \forall j, k \in \mathrm{D}. 
\end{align}
We therefore define the maximum accuracy disparity of a classifier as follows:
\begin{align}
\epsilon (\widehat{Y}) = \max_{\forall j,k \in \mathrm{D}} \bigg( \log \frac{P(\widehat{Y} = Y | A = j)}{P(\widehat{Y} = Y | A = k)} \bigg).
\end{align}

Table~\ref{tab:genrace} shows the gender classification accuracy of different models measured on the external validation datasets for each race and gender group. The FairFace model achieves the lowest maximum accuracy disparity. The LFWA+ model yields the highest disparity, strongly biased toward the male cateogory. The CelebA model tends to exhibit a bias toward the female category as the dataset contains more female images than male. 

The FairFace model achieves less than 1\% accuracy discrepancy between male $\leftrightarrow$  female and White $\leftrightarrow$ non-White for gender classification (Table~\ref{table:validation}). All the other models show a strong bias toward the male class, yielding much lower accuracy on the female group, and perform more inaccurately on the non-White group. The gender performance gap was the biggest in LFWA+ (32\%), which is the smallest among the datasets used in the experiment. Recent work has also reported asymmetric gender biases in commercial computer vision services~\cite{buolamwini2018gender}, and our result further suggests the cause is likely to due to the unbalanced representation in training data. 

\begin{table*}[]
\resizebox{\textwidth}{!}{%
\begin{tabular}{|c|cccccccccccccc|ccccc}
\cline{1-15}
Race     & \multicolumn{2}{c}{White} & \multicolumn{2}{c}{Black} & \multicolumn{2}{c}{East Asian} & \multicolumn{2}{c}{SE Asian} & \multicolumn{2}{c}{Latino} & \multicolumn{2}{c}{Indian} & \multicolumn{2}{c|}{Middle Eastern} &               &               &               &               &                                    \\ \hline
Gender   & M           & F           & M           & F           & M              & F             & M             & F            & M            & F           & M            & F           & M                & F                & Max           & Min           & AVG           & STDV          & \multicolumn{1}{c|}{$\epsilon$}             \\ \hline
FairFace & .967        & .954        & .958        & .917        & .873           & .939          & .909          & .906         & .977         & .960        & .966         & .947        & .991             & .946             & \textbf{.991} & \textbf{.873} & \textbf{.944} & \textbf{.032} & \multicolumn{1}{c|}{\textbf{.055}} \\
UTK      & .926        & .864        & .909        & .795        & .841           & .824          & .906          & .795         & .939         & .821        & .978         & .742        & .949             & .730             & .978          & .730          & .859          & .078          & \multicolumn{1}{c|}{.127}          \\
LFWA+    & .946        & .680        & .974        & .432        & .826           & .684          & .938          & .574         & .951         & .613        & .968         & .518        & .988             & .635             & .988          & .432          & .766          & .196          & \multicolumn{1}{c|}{.359}          \\
CelebA   & .829        & .958        & .819        & .919        & .653           & .939          & .768          & .923         & .843         & .955        & .866         & .856        & .924             & .874             & .958          & .653          & .866          & .083          & \multicolumn{1}{c|}{.166}          \\ \hline
\end{tabular}%
}
\caption{Gender classification accuracy measured on external validation datasets across gender-race groups. }
\label{tab:genrace}
\end{table*}


\textbf{Data Coverage and Diversity. }
We further investigate dataset characteristics to measure the data diversity in our dataset. We first visualize randomly sampled faces in 2D space using t-SNE \cite{maaten2008visualizing} as shown in Figure~\ref{fig:tsne}. We used the facial embedding based on ResNet-34 from dlib, which was trained from the FaceScrub dataset~\cite{ng2014data}, the VGG-Face dataset~\cite{parkhi2015deep} and other online sources, which are likely dominated by the White faces. The faces in FairFace are well spread in the space, and the race groups are loosely separated from each other. This is in part because the embedding was trained from biased datasets, but it also suggests that the dataset contains many non-typical examples. 
LFWA+ was derived from LFW, which was developed for face recognition, and therefore contains multiple images of the same individuals, \ie, clusters. UTKFace also tends to focus more on local clusters compared to FairFace.



\begin{figure}
  \centering
      \includegraphics[width=0.4\textwidth]{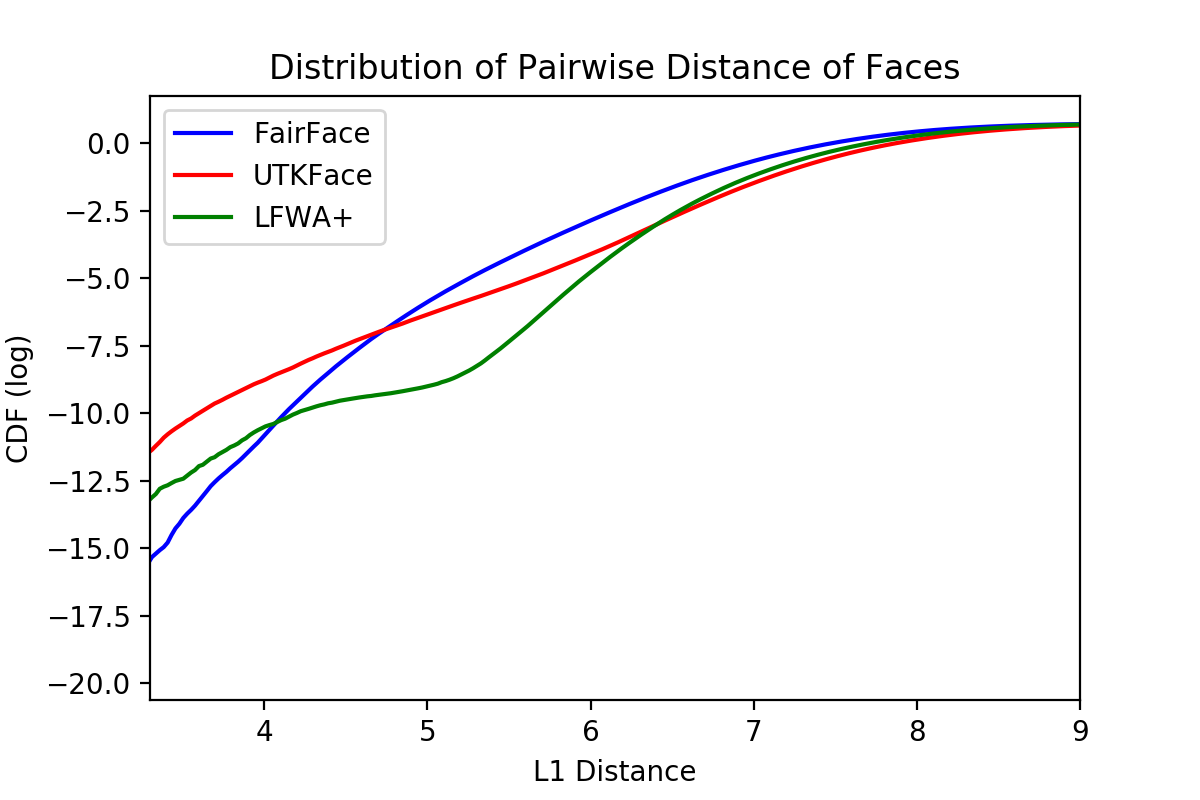}
  \caption{Distribution of pairwise distances of face in 3 datasets measured by L1 distance on face embedding. }
\label{fig:cdf}
\end{figure}

To explicitly measure the diversity of faces in these datasets, we examine the distributions of pairwise distance between faces (Figure~\ref{fig:cdf}). On the random subsets, we first obtained the same 128-dimensional facial embedding from dlib and measured pair-wise distance. Figure~\ref{fig:cdf} shows the CDF functions for 3 datasets. As conjectured, UTKFace had more faces that are tightly clustered together and very similar to each other, compared to our dataset. Surprisingly, the faces in LFWA+ were shown very diverse and far from each other, even though the majority of the examples contained a white face. We believe this is mostly due to the fact that the face embedding was also trained on a very similar white-oriented dataset which will be effective in separating white faces, not because the appearance of their faces is actually diverse. (See~Figure~\ref{fig:grid})

\begin{table*}
\small
\caption{Cross-Dataset Classification Accuracy on White Race.}
\label{table:cross_white}
\vspace{-10pt}
\centering
\scalebox{1.0}{
\begin{tabular}{|c|c|c|c|c|c|c|c|c|c|c|}
\hline
\multirow{3}{*}{}           & \multicolumn{10}{c|}{Tested on}                                                                                                                                       \\ \cline{2-11} 
                            & \multicolumn{4}{c|}{Race}                                   & \multicolumn{4}{c|}{Gender}                                           & \multicolumn{2}{c|}{Age}        \\ \cline{2-11} 
                            &          & FairFace       & UTKFace        & LFWA+          & FairFace       & UTKFace        & LFWA+          & CelebA* & FairFace       & UTKFace        \\ \hline
\multirow{4}{*}{Trained on} & FairFace & \textbf{.937} & .936 & \textbf{.970} & \textbf{.942} & \textbf{.940} & .920          & \textbf{.981}     & \textbf{.597} & .565          \\ \cline{2-11} 
                            & UTKFace  & .800          & .918          & .925          & .860          & .935          & .916          & .962              & .413          & \textbf{.576} \\ \cline{2-11} 
                            & LFWA+    & .879          & \textbf{.947}          & .961          & .761          & .842          & \textbf{.930} & .940              & -              & -              \\ \cline{2-11} 
                            & CelebA   & -             & -             & -             & .812          & .880          & .905          & .971              & -              & -              \\ \hline
                            \multicolumn{11}{r}{* CelebA doesn't provide race annotations. The result was obtained from the whole set (white and non-white). } 
\end{tabular}
}
\end{table*}


\begin{table*}
\small
\caption{Cross-Dataset Classification Accuracy on non-White Races.}
\vspace{-10pt}
\label{table:cross_non_white}
\centering
\scalebox{1.0}{
\begin{tabular}{|c|c|c|c|c|c|c|c|c|c|c|}
\hline
\multirow{3}{*}{}           & \multicolumn{10}{c|}{Tested on}                                                                                                                                       \\ \cline{2-11} 
                            & \multicolumn{4}{c|}{Race$\dagger$}                                   & \multicolumn{4}{c|}{Gender}                                           & \multicolumn{2}{c|}{Age}        \\ \cline{2-11} 
                            &          & FairFace       & UTKFace        & LFWA+          & FairFace       & UTKFace        & LFWA+          & CelebA* & FairFace       & UTKFace        \\ \hline
\multirow{4}{*}{Trained on} & FairFace & \textbf{.754} & .801          & \textbf{.960} & \textbf{.944} & \textbf{.939} & \textbf{.930} & \textbf{.981}     & \textbf{.607} & .616   \\ \cline{2-11} 
                            & UTKFace  & .693          & \textbf{.839} & .887          & .823          & .925          & .908          & .962              & .418          & \textbf{.617}            \\ \cline{2-11} 
                            & LFWA+    & .541          & .380          & .866          & .738          & .833          & .894          & .940              & -              & -              \\ \cline{2-11} 
                            & CelebA   & -             & -             & -             & .781          & .886          & .901          & .971     & -              & -              \\ \hline
                            \multicolumn{11}{r}{* CelebA doesn't provide race annotations. The result was obtained from the whole set (white and non-white). } \\
                            \multicolumn{11}{r}{$\dagger$ FairFace defines 7 race categories but only 4 races (White, Black, Asian, and Indian) were used in this result }         \\
                            \multicolumn{11}{r}{to make it comparable to UTKFace. }                       
\end{tabular}  
}
\end{table*}

\begin{table*}[]
\small
\caption{Gender classification accuracy on external validation datasets, across race and age groups.}
\vspace{-10pt}
\label{table:mean_var}
\centering
\scalebox{1.0}{
\begin{tabular}{|c|c|c|c|c|c|}
\hline
\multicolumn{2}{|c|}{}                                                                  & Mean across races & SD across races & Mean across ages & SD across ages \\ \hline
\multirow{4}{*}{\begin{tabular}[c]{@{}c@{}}Model\\  trained on\end{tabular}} & FairFace & \textbf{94.89\%}  & \textbf{3.03\%} & \textbf{92.95\%} & \textbf{6.63\%}      \\ \cline{2-6} 
                                                                             & UTKFace  & 89.54\%           & 3.34\%          & 84.23\%          & 12.83\%               \\ \cline{2-6} 
                                                                             & LFWA+    & 82.46\%           & 5.60\%          & 78.50\%          & 11.51\%               \\ \cline{2-6} 
                                                                             & CelebA   & 86.03\%           & 4.57\%          & 79.53\%          & 17.96\%              \\ \hline
\end{tabular}
}
\end{table*}


%% file: 060-bigtable.tex
\begin{table*}
\centering
\caption{Classification accuracy on external validation datasets.}
\label{table:validation}
\centering
\scalebox{0.72}{
\begin{tabular}{|c|c||c||c|c|c|c|c|c|c|c|c|c|c|c|c|c|c|}
\hline
\multicolumn{2}{|c||}{\multirow{2}{*}{}} & \multicolumn{16}{c|}{Race Classification}                                                           \\ \cline{3-18} 
\multicolumn{2}{|c||}{}                  & All           & Female        & Male          & White         & {\footnotesize Non-White}     & Black         & Asian         & {\footnotesize E Asian}    & {\footnotesize SE Asian}      & Latino        & Indian        & {\footnotesize Mid-East}   & 0-9           & 10-29         & 30-49         & 50+ \\ \hline
\multirow{3}{*}{Twitter}   & FairFace   & \textbf{.733} & \textbf{.726} & \textbf{.737} & .899          & \textbf{.548} & \textbf{.695} & \textbf{.888} & .705          & .465          & .305          & \textbf{.492}          & .743          & \textbf{.756} & \textbf{.691}          & \textbf{.768} & \textbf{.777}     \\ \cline{2-18} 
                           & UTKFace    & .544          & .543          & .544          & .741          & .354          & .591          & .476          & -             & -             & -             & .474 & -             & .606          & .516          & .574          & .567              \\ \cline{2-18} 
                           & LFWA+      & .626          & .596          & .647          & \textbf{.965} & .284          & .283          & .425          & -             & -             & -             & -             & -             & .639          & .562 & .705          & .751              \\ \hline
\multirow{3}{*}{Media}     & FairFace   & \textbf{.866} & \textbf{.874} & \textbf{.863} & .949          & \textbf{.685} & \textbf{.890} & \textbf{.918} & .886          & .152          & .267          & \textbf{.691} & .704          & \textbf{.833} & \textbf{.853} & \textbf{.852} & \textbf{.893}     \\ \cline{2-18} 
                           & UTKFace    & .772          & .795          & .763          & .883          & .546          & .802          & .588          & -             & -             & -             & .599          & -             & .646          & .755          & .757          & .804              \\ \cline{2-18} 
                           & LFWA+      & .679          & .823          & .835          & \textbf{.978} & .393          & .485          & .578          & -             & -             & -             & -             & -             & .682          & .656          & .651          & .722              \\ \hline
\multirow{3}{*}{Protest}   & FairFace   & \textbf{.846} & \textbf{.849} & \textbf{.844} & .935          & \textbf{.683} & \textbf{.859} & \textbf{.843} & .702          & .510          & .169          & \textbf{.649}          & .779          & \textbf{.839} & \textbf{.821} & \textbf{.837} & \textbf{.881}     \\ \cline{2-18} 
                           & UTKFace    & .706          & .723          & .697          & .821          & .536          & .714          & .456          & -             & -             & -             & .591 & -             & .681          & .658          & .685          & .787              \\ \cline{2-18} 
                           & LFWA+      & .747          & .759          & .741          & \textbf{.964} & .366          & .418          & .488          & -             & -             & -             & -             & -             & .689          & .645          & .668          & .801              \\ \hline
\multirow{5}{*}{Average}   & FairFace   & \textbf{.815} & \textbf{.816} & \textbf{.815} & .928          & \textbf{.639} & \textbf{.815} & \textbf{.883} & .764          & .376          & .247          & \textbf{.611}          & .742          & \textbf{.809} & \textbf{.788} & \textbf{.819} & \textbf{.850}     \\ \cline{2-18} 
                           & FairFace 18K & .800	& .812	& .795	& .917	& .588	& .779	& .856	& .685	& .355	& .279	& .502	& .625	& .786	& .773	& .809	& .827
              \\ \cline{2-18} 
                           & FairFace 9K & .774	& .788	& .768	& .885	& .564	& .756	& .827	& .641	& .315	& .281	& .531	& .544	& .723	& .757	& .789	& .787
             \\ \cline{2-18} 
                           & UTKFace    & .674          & .687          & .668          & .815          & .479          & .702          & .507          & -             & -             & -             & .555 & -             & .644          & .643          & .672          & .719              \\ \cline{2-18} 
                           & LFWA+      & .684          & .726          & .741          & \textbf{.969} & .348          & .395          & .497          & -             & -             & -             & -             & -             & .670          & .621          & .675          & .758              \\ \hline
\multicolumn{2}{|c|}{\multirow{2}{*}{}} & \multicolumn{16}{c|}{Gender Classification}                                                        \\ \cline{3-18} 
\multicolumn{2}{|c|}{}                  & All           & Female        & Male          & White         & {\footnotesize Non-White}    & Black         & Asian         & {\footnotesize E Asian}    & {\footnotesize SE Asian}      & Latino        & Indian        & {\footnotesize Mid-East}   & 0-9           & 10-29         & 30-49         & 50+ \\ \hline
\multirow{4}{*}{Twitter}   & FairFace   & \textbf{.940} & .948 & \textbf{.935} & \textbf{.949} & \textbf{.932} & \textbf{.932} & \textbf{.894} & \textbf{.864} & \textbf{.942} & \textbf{.963} & \textbf{.932} & \textbf{.976} & \textbf{.817} & \textbf{.932} & \textbf{.973} & \textbf{.959}     \\ \cline{2-18} 
                           & UTKFace    & .884          & .859          & .899          & .897          & .874          & .864          & .829          & .803          & .871          & .901          & .898          & .947          & .671          & .874          & .933          & .912              \\ \cline{2-18} 
                           & LFWA+      & .797          & .637          & .899          & .815          & .773          & .789          & .724          & .716          & .736          & .804          & .728          & .911          & .634          & .769          & .857          & .859              \\ \cline{2-18} 
                           & CelebA     & .829          & \textbf{.955}          & .750          & .850          & .812          & .818          & .764          & .716          & .839          & .843          & .831          & .876          & .539          & .818          & .889          & .881              \\ \hline
\multirow{4}{*}{Media}     & FairFace   & \textbf{.973} & \textbf{.957} & \textbf{.980} & \textbf{.976} & \textbf{.969} & \textbf{.953} & \textbf{.956} & \textbf{.967} & \textbf{.891} & \textbf{.980} & \textbf{.977} & \textbf{.988} & \textbf{.821} & \textbf{.952} & \textbf{.984}          & \textbf{.979}     \\ \cline{2-18} 
                           & UTKFace    & .927          & .841          & .961          & .928          & .915          & .907          & .908          & .915          & .869          & .928          & .945          & .932          & .679          & .917          & .931          & .924              \\ \cline{2-18} 
                           & LFWA+      & .887          & .656          & .976          & .893          & .871          & .851          & .864          & .875          & .804          & .859          & .897          & .944          & .688          & .835          & .832          & .911              \\ \cline{2-18} 
                           & CelebA     & .899          & .950          & .880          & .909          & .881          & .847          & .858          & .857          & .870          & .925          & .884          & .926          & .560          & .860          & .908 & .924              \\ \hline
\multirow{4}{*}{Protest}   & FairFace   & \textbf{.957} & \textbf{.944} & \textbf{.963} & \textbf{.962} & \textbf{.951} & \textbf{.957} & .887 & \textbf{.879} & \textbf{.906} & \textbf{.970} & \textbf{.973} & \textbf{.991} & \textbf{.861} & \textbf{.934} & \textbf{.967} & \textbf{.976}     \\ \cline{2-18} 
                           & UTKFace    & .901          & .829          & .934          & .905          & .873          & .911          & .814          & .802          & .843          & .902          & .918          & .921          & .611          & .812          & .924          & .919              \\ \cline{2-18} 
                           & LFWA+      & .829          & .567          & .954         & .841          & .801          & .821          & .758          & .782          & .697          & .811          & .811          & .929          & .568          & .705          & .851          & .908              \\ \cline{2-18} 
                           & CelebA     & .882          & .935          & .856          & .893          & .866          & .876          & \textbf{.892}          & .750          & .833          & .892          & .878          & .956          & .492          & .842          & .904          & .927              \\ \hline
\multirow{6}{*}{Average}   & FairFace   & \textbf{.957} & \textbf{.950} & \textbf{.959} & \textbf{.962} & \textbf{.951} & \textbf{.947} & \textbf{.912} & \textbf{.903} & \textbf{.913} & \textbf{.971} & \textbf{.961} & \textbf{.985} & \textbf{.833} & \textbf{.939} & \textbf{.975} & \textbf{.971}     \\ \cline{2-18} 
                           & FairFace 18K & .941	 & .930	 & .946	 & .946	 & .934	 & .931	 & .891	 & .886	 & .895	 & .955	 & .960	 & .967	 & .803	 & .920	 & .957	 & .962
              \\ \cline{2-18}
                           & FairFace 9K & .926	 & .921	 & .927	 & .929	 & .921	 & .922	 & .864	 & .851	 & .883	 & .942 & .951	 & .974	 & .760	 & .901	 & .949	 & .943
              \\ \cline{2-18}
                           & UTKFace    & .904          & .843          & .931          & .910          & .887          & .894          & .850          & .840          & .861          & .910          & .920          & .933          & .654          & .868          & .929          & .918              \\ \cline{2-18} 
                           & LFWA+      & .838          & .620          & .943          & .850          & .815          & .820          & .782          & .791          & .746          & .825          & .812          & .928          & .630          & .770          & .847          & .893              \\ \cline{2-18} 
                           & CelebA     & .870          & .947          & .829          & .884          & .853          & .847          & .838          & .774          & .847          & .887          & .864          & .919          & .530          & .840          & .900          & .911              \\ \hline
\multicolumn{2}{|c|}{\multirow{2}{*}{}} & \multicolumn{16}{c|}{Age Classification}                                                                                                                                                                                                                                         \\ \cline{3-18} 
\multicolumn{2}{|c|}{}                  & All           & Female        & Male          & White         & {\footnotesize Non-White}     & Black         & Asian         & {\footnotesize E Asian}    & {\footnotesize SE Asian}      & Latino        & Indian        & {\footnotesize Mid-East}   & 0-9           & 10-29         & 30-49         & 50+ \\ \hline
\multirow{2}{*}{Twitter}   & FairFace   & \textbf{.578} & \textbf{.586} & \textbf{.573} & \textbf{.563} & \textbf{.590} & \textbf{.557} & \textbf{.620} & \textbf{.629} & \textbf{.606} & \textbf{.581} & \textbf{.576} & \textbf{.555} & \textbf{.805} & \textbf{.666} & \textbf{.439} & \textbf{.408}     \\ \cline{2-18} 
                           & UTKFace    & .366          & .355          & .384          & .343          & .385          & .338          & .397          & .382          & .419          & .411          & .356          & .345          & .585          & .499          & .104          & .307              \\ \hline
\multirow{2}{*}{Media}     & FairFace   & \textbf{.516} & \textbf{.511} & \textbf{.517} & \textbf{.513} & \textbf{.520} & \textbf{.483} & \textbf{.557} & \textbf{.559} & \textbf{.543} & \textbf{.537} & \textbf{.532} & \textbf{.475} & \textbf{.714} & \textbf{.686} & \textbf{.447} & \textbf{.501}     \\ \cline{2-18} 
                           & UTKFace    & .275          & .273          & .282          & .281          & .267          & .271          & .276          & .279          & .261          & .231          & .292          & .222          & .511          & .529          & .112          & .238              \\ \hline
\multirow{2}{*}{Protest}   & FairFace   & \textbf{.515} & \textbf{.543} & \textbf{.502} & \textbf{.498} & \textbf{.539} & \textbf{.527} & \textbf{.584} & \textbf{.605} & \textbf{.531} & \textbf{.507} & \textbf{.581} & \textbf{.469} & \textbf{.885} & \textbf{.687} & \textbf{.395} & \textbf{.478}     \\ \cline{2-18} 
                           & UTKFace    & .302          & .306          & .294          & .291          & .319          & .305          & .316          & .318          & .312          & .314          & .371          & .318          & .516          & .503          & .114          & .349              \\ \hline
\multirow{4}{*}{Average}   & FairFace   & \textbf{.536} & \textbf{.547} & \textbf{.531} & \textbf{.525} & \textbf{.550} & \textbf{.522} & \textbf{.587} & \textbf{.598} & \textbf{.560} & \textbf{.542} & \textbf{.563} & \textbf{.500} & \textbf{.801} & \textbf{.680} & \textbf{.427} & \textbf{.462}     \\ \cline{2-18} 
                           & FairFace 18K & .492	& .508	& .484	& .485	& .496	& .463	& .528	& .538	& .506	& .510	& .454	& .490	& .700	& .646	& .387	& .410
              \\ \cline{2-18}
                           & FairFace 9K & .470	& .493	& .459	& .462	& .478	& .449	& .506	& .515	& .483	& .473	& .458	& .463	& .662	& .611	& .361	& .394
              \\ \cline{2-18}
                           & UTKFace    & .314          & .311          & .320          & .305          & .324          & .305          & .330          & .326          & .331          & .319          & .340          & .295          & .537          & .510          & .110          & .298              \\ \hline
\end{tabular}
}
\end{table*}


%% file: 050-conclusion.tex
\section{Conclusion}
This paper proposes a novel face image dataset balanced on race, gender and age. Compared to existing large-scale in-the-wild datasets, our dataset achieves much better generalization classification performance for gender, race, and age on novel image datasets collected from Twitter, international online newspapers, and web search, which contain more non-White faces than typical face datasets. 
We show that the model trained from our dataset produces balanced accuracy across race, whereas other datasets often lead to asymmetric accuracy on different race groups. 

This dataset was derived from the Yahoo YFCC100m dataset~\cite{thomee59yfcc100m} for the images with Creative Common Licenses by Attribution and Share Alike, which permit both academic and commercial usage. Our dataset can be used for training a new model and verifying balanced accuracy of existing classifiers. 


Algorithmic fairness is an important aspect to consider in designing and developing AI systems, especially because these systems are being translated into many areas in our society and affecting our decision making. Large scale image datasets have contributed to the recent success in computer vision by improving model accuracy; yet the public and media have doubts about its transparency. The novel dataset proposed in this paper will help us discover and mitigate race and gender bias present in computer vision systems such that such systems can be more easily accepted in society. 

